\documentclass[accepted]{uai2026} 
                        
\usepackage[american]{babel}

\usepackage{natbib} 
\usepackage{mathtools} 
\usepackage{booktabs} 
\usepackage{tikz} 
\usepackage{amssymb}
\usepackage{adjustbox}
\usepackage{multicol}
\usepackage{multirow}
\usepackage{placeins}

\title{APIC: Amortized Physics-Informed Calibration using Neural Processes}

\author[1]{\href{mailto:aishwarya.venkataramanan@uni-jena.de}
{Aishwarya Venkataramanan$^{*}$}}

\author[1]{\href{mailto:sai.karthikeya.vemuri@uni-jena.de}
{Sai Karthikeya Vemuri$^{*}$}}

\author[1]{\href{mailto:joachim.denzler@uni-jena.de}
{Joachim Denzler}}

\affil[1]{Computer Vision Group, Friedrich Schiller University Jena, Germany}

\begin{document}
\maketitle
\begingroup
\renewcommand\thefootnote{}
\footnotetext{* Equal contribution.}
\endgroup

\begin{abstract}

Physics models are inherently imperfect due to misspecified or missing mechanisms, resulting in systematic discrepancies between model predictions and real-world observations. The Kennedy–O’Hagan (KOH) framework addresses this issue through explicit discrepancy modeling. However, its non-amortized, per-instance formulation limits scalability across families of related systems. We introduce Amortized Physics-Informed Calibration (APIC), a population-level extension of KOH that leverages Neural Processes to perform scalable Bayesian inference across realizations. Our framework employs a two-branch latent architecture to disentangle instance-specific physical parameters from shared, state-dependent structural discrepancies. By integrating differentiable physics into an amortized inference backbone, APIC enables rapid calibration of unseen realizations from sparse observations while quantifying uncertainty. Experiments on the damped spring oscillator, the Lotka–Volterra system, and the advection–diffusion PDE with misspecified physics demonstrate improved parameter recovery and consistent identification of the systemic discrepancy structure compared to other calibration approaches.

\end{abstract}

\section{Introduction}\label{sec:intro}

Physical simulators are ubiquitous in scientific modeling, enabling the prediction of complex phenomena ranging from climate dynamics~\citep{lai2025machine} to aerostructural mechanics~\citep{catalani2024neural}. Yet even high-fidelity computational models inevitably exhibit systematic discrepancies with real-world observations due to unresolved physics, modeling assumptions, or numerical approximations~\citep{KOH2001}. In practice, observed system behavior reflects two distinct sources of uncertainty. The first is parametric uncertainty, corresponding to the “known unknowns” of the model: the parameters explicitly represented in the simulator’s equations but whose numerical values are unknown. The second is structural discrepancy, representing the “unknown unknowns”: systematic deviations arising from model inadequacy, where relevant physical processes or dependencies are absent from the governing equations altogether. Disentangling these two sources of variation is generally non-identifiable without additional structural assumptions. During model calibration, i.e., adjusting parameters to fit observed data, structural errors can mimic or compensate for parameter misspecification, and conversely, parameter adjustments can absorb unmodeled discrepancy. In such cases, the optimizer may find parameters that minimize the residual error but lead to non-identifiability and compromise physical interpretability~\citep{jenny2014}.

The Kennedy–O’Hagan (KOH)~\citep{KOH2001} framework is a seminal approach for disentangling parametric uncertainty from structural discrepancy in physical models. It explicitly decomposes observed data $\mathbf{y}$ into three components: a physical simulator $\mathcal{F}$, a systematic discrepancy term, and observational noise, formalized as

\begin{equation}
    \mathbf{y} = \mathcal{F}(\boldsymbol{\theta}) + \boldsymbol{\delta} + \boldsymbol{\epsilon},
\end{equation}
where $\boldsymbol{\theta}$ represents the optimal (but unknown) physical parameters, $\boldsymbol{\delta}$ captures systematic model discrepancy, and $\boldsymbol{\epsilon}$ represents observational noise. Here, $\boldsymbol{\delta}$ is intended to capture only variability that cannot be explained by the simulator. If the discrepancy is sufficiently flexible to reproduce the simulator’s dynamics, it can compensate for misspecified physical parameters. This results in non-identifiability between $\boldsymbol{\delta}$ and $\boldsymbol{\theta}$ and thus compromises physical interpretability.

KOH employs Gaussian Processes \citep{Rasmussen2006Gaussian} to simultaneously estimate unknown parameters and learn the discrepancy. Despite its theoretical elegance, the classical framework is non-amortized. 
For each new realization (e.g., a new experiment or system instance), inference over $\boldsymbol{\theta}$ and $\boldsymbol{\delta}$ must be performed from scratch. 
This paradigm does not exploit the fact that although parameters may vary across instances, the underlying physical mechanisms that govern the discrepancy are often shared across a population.

In parallel, the development of hybrid modeling has aimed to bridge the gap between purely physical simulators and data-driven approaches by integrating physical laws with deep learning~\citep{reichstein2019deep, benson2025atmospheric}. Such approaches include Physics-Informed Neural Networks (PINNs) \citep{Raissi2017}, residual learning frameworks \citep{pmlr-v151-wang22a, takeishi2021physics}, and neural operator-based corrections \citep{Li2020FourierNO}, which correct imperfect simulators using observed data. 
While these methods often improve predictive accuracy, they primarily optimize pointwise reconstruction objectives and do not typically adopt a fully probabilistic formulation. As a result, uncertainty quantification is often limited or heuristic, and identifiability between physical parameters and learned discrepancy remains insufficiently addressed.

\begin{figure*}
    \centering
    \includegraphics[width=\linewidth]{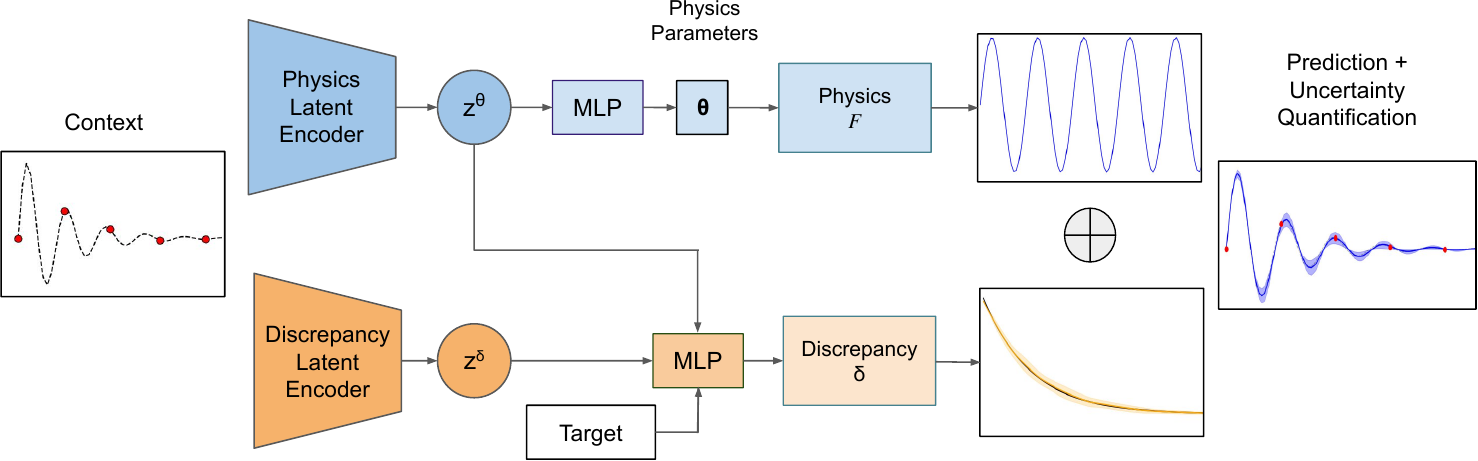}
    \caption{\textbf{Overview of Amortized Physics-Informed Calibration (APIC)}: The framework employs a dual-branch latent-variable Neural Process architecture to disentangle observed system behavior into parametric physics and a structural discrepancy. The \textit{Physics Branch} (top) infers instance-specific parameters $\theta$ to drive a differentiable simulator $F$, while the \textit{Discrepancy Branch} (bottom) learns a shared, state-dependent correction model $\delta$. By combining these components, APIC achieves calibration and uncertainty quantification for unseen realizations of imperfect physical systems.}
    \label{fig:apic}
\end{figure*}

We propose \textbf{\emph{Amortized Physics-Informed Calibration (APIC)}}: a framework that unites the probabilistic rigor of KOH with the amortized efficiency of deep generative models \citep{jha2023neuralprocessfamilysurvey}. APIC models a population of related system realizations as arising from a shared physical simulator and a learned, state-dependent correction for structured discrepancy over the input domain.
The model consists of a dual-latent encoder architecture. 
One latent branch captures instance-specific physical parameters, which are mapped to the simulator’s input. A second latent branch encodes residual discrepancies. Given sparse observations from a new realization, APIC infers both latent variables, combines the simulator output with the discrepancy network, and produces predictive distributions with principled uncertainty estimates. To encourage meaningful separation between the physics and discrepancy components, the training proceeds in two stages. We first learn the mapping between observations and physical parameters using simulator-generated data with known ground-truth parameters. 
We then train on real observations while regularizing the discrepancy component to capture systematic errors without undermining physical interpretability. The model is presented in \autoref{fig:apic}. 

We evaluate APIC on three representative dynamical systems: a damped spring oscillator, a predator-prey ordinary differential equation (ODE) model, and an advection-diffusion system exhibiting systematic discrepancies. 
Across the systems, we assess reconstruction accuracy, predictive uncertainty, and the accuracy of parameter and discrepancy recovery. 
Our results show that APIC achieves improved predictive performance while maintaining accurate and physically interpretable parameter recovery.

In summary, our contributions are threefold:
(1) We introduce APIC, a probabilistic framework that integrates physics-based simulation with amortized latent-variable inference for population-level calibration.
(2) We propose a dual-latent architecture that disentangles instance-specific physical parameters from structured model discrepancies, enabling physically interpretable inference under sparse observations.
(3) We demonstrate that APIC achieves improved predictive performance and reliable uncertainty quantification across multiple dynamical systems. The code is available in: \url{https://github.com/cvjena/APIC.git}.

\section{Related Works}

The foundational framework for computer model calibration was established by Kennedy and O’Hagan \cite{KOH2001}, who formulated calibration as a decomposition of observations into a parametric simulator output and a systematic discrepancy term representing missing physics. While the KOH framework provides a basis for uncertainty quantification, it suffers from two major limitations: identifiability, where the discrepancy term confounds inference of physically meaningful parameters \citep{jenny2014}, and computational intractability. Classical implementations rely on Gaussian Processes (GPs) and Markov Chain Monte Carlo (MCMC), which scale cubically with data size and require full re-inference for each new experimental realization \citep{Higdon2004}. Subsequent work introduced orthogonality constraints to mitigate identifiability issues \citep{Bayyari2007,Tuo2015,Plumlee03072017}, but per-instance inference remains a computational bottleneck for large-scale or real-time applications.

Recent advances in deep learning, particularly Physics-Informed Neural Networks (PINNs) \citep{Raissi2017,vemuri2024}, have shifted physics modeling toward gradient-based optimization by embedding physical residuals directly into the loss function. Extensions have incorporated uncertainty modeling and structural error correction. For example, Bayesian PINNs (B-PINNs) \citep{YANG2021109913} quantify uncertainty in inverse problems, while \citep{ZOU2024112918,pmlr-v151-wang22a,BCPI} introduce secondary networks or Deep Gaussian Processes to account for misspecified differential equations. Although effective, these methods remain optimization-based: they solve a new inverse problem for each dataset and do not learn an amortized mapping from observations to parameters. This limits their scalability in settings requiring rapid calibration across many realizations.

Neural Processes (NPs) \citep{garnelo2018conditional, garnelo2018neural} provide a probabilistic, amortized framework for learning distributions over functions from sparse context observations. By encoding context–target relationships into latent variables, NPs enable fast adaptation to new tasks without per-task optimization. Subsequent extensions~\citep{kim2019attentive, venkataramanan2025distance} improve expressiveness and uncertainty calibration. However, standard NPs are purely data-driven: they do not incorporate known physical simulators or account for structured model discrepancies.

Amortized approaches \citep{KyleCramner2020,Pedro2020} learn direct mappings from data to posterior distributions, shifting computational cost to an offline training phase and enabling fast inference. Neural operator methods, such as DeepONet \citep{Lu2021} and Fourier Neural Operators \citep{Li2020FourierNO}, further enable resolution-independent mappings between function spaces. While promising, these methods generally treat the underlying physics either as a black box (ignoring known physical constraints) or as a perfect white box (ignoring structural discrepancies).

To the best of our knowledge, there is currently no unified framework that integrates deep learning-based amortized inference with the KOH calibration principle. 
We address this gap by introducing an amortized, physics-informed Neural Process that explicitly disentangles physical parameter inference from structured discrepancy modeling across a family of related system realizations. This formulation preserves physical interpretability while enabling efficient inference and principled uncertainty quantification.

\section{Method}\label{sec:method}

We present Amortized Physics-Informed Calibration (APIC), a population-level extension of the KOH framework that generalizes single-instance Bayesian calibration to families of related tasks. The method employs a dual-encoder Neural Process architecture to model realization-specific physical parameters and structural discrepancy, together with a two-stage curriculum training strategy that stabilizes learning and promotes their disentanglement.

\subsection{Problem Formulation: Meta-Level Calibration}

We extend the classical KOH framework to a family of related
realizations $\mathcal{Y} = \{\mathbf{y}_i\}_{i=1}^N$. Observations
for each realization are modeled as
\begin{equation}
    \mathbf{y}_i =
    \mathcal{F}(\mathbf{x}_i; \boldsymbol{\theta}_i)
    \;\circledast\;
    \boldsymbol{\delta}(\mathbf{x}_i; \boldsymbol{\theta}_i)
    + \boldsymbol{\epsilon}_i,
\end{equation}
where $\circledast$ denotes an operator that specifies how structural model error modifies the nominal solver output. Classical KOH corresponds to the additive case, while multiplicative corrections are also accommodated in our formulation.
$\mathcal{F}$ is a differentiable physics solver, $\mathbf{x}_i$ denotes the system state (e.g., time or observable variables),
$\boldsymbol{\theta}_i$ are realization-specific physical parameters, and
$\boldsymbol{\epsilon}_i$ represents observation noise.
Our objective is to jointly infer the latent physical parameters
$\boldsymbol{\theta}_i$ and the corresponding discrepancy functions
$\boldsymbol{\delta}(\mathbf{x}_i; \boldsymbol{\theta}_i)$ across realizations.

To this end, we employ a Neural Process framework to amortize inference over the family of realizations. This enables mapping sparse observations to posterior distributions over latent variables governing both physical parameters and structured discrepancy. The model is illustrated in \autoref{fig:apic}.

\subsection{APIC model Architecture}

For each realization $i$, we observe a set of $T_i$ state-observation pairs $\{(\mathbf{x}_{ij}, \mathbf{y}_{ij})\}_{j=1}^{T_i}$.
Following the NP framework, we partition these observations into a context set $\mathcal{C}_i = \{(\mathbf{x}_{ij}^{(c)}, \mathbf{y}_{ij}^{(c)})\}_{j=1}^{C_i}$
and a target set $\mathcal{T}_i = \{(\mathbf{x}_{ij}^{(t)}, \mathbf{y}_{ij}^{(t)})\}_{j=1}^{T_i}$, such that $\mathcal{C}_i \subseteq \mathcal{T}_i$. The context set provides sparse observations
used for inference, while the target set specifies the states at which predictions are required.

The encoder consists of 2 latent inference networks that operate on the context set $\mathcal{C}_i$. The physics encoder infers a latent representation
$\mathbf{z}_i^{(\theta)}$ that captures instance-specific physical properties. This is further mapped to the solver parameters
$\boldsymbol{\theta}_i$.
In parallel, a discrepancy encoder infers a latent variable
$\mathbf{z}_i^{(\delta)}$ that parameterizes a state-dependent discrepancy function.
The decoder combines the inferred physical parameters, discrepancy latent, and target states to produce predictive distributions at the target inputs.

\paragraph{Physics Latent Encoder.}
The physics latent encoder infers a latent representation of the instance-specific
physical parameters from the context set $\mathcal{C}_i$. Each context pair
$(\mathbf{x}_{ij}^{(c)}, \mathbf{y}_{ij}^{(c)})$ is mapped through a shared
encoder network $h_\theta(\cdot)$ to obtain a representation. The individual
context representations are aggregated into a single latent embedding that
summarizes information from the context set. 
Depending on the specific NP variant, this aggregation can be a simple mean or sum over embeddings, or a learned attention mechanism.
This aggregated representation is used to produce a Gaussian posterior over the physics latent variable:

\begin{equation}
q\!\left(\mathbf{z}_i^{(\theta)} \mid \mathcal{C}_i\right)
= \mathcal{N}\!\left(\boldsymbol{\mu}_\theta(\mathcal{C}_i), \boldsymbol{\Sigma}_\theta(\mathcal{C}_i)\right),
\end{equation}
where $\boldsymbol{\mu}_\theta(\cdot)$ and $\boldsymbol{\Sigma}_\theta(\cdot)$
are neural networks producing the mean and diagonal covariance, respectively.

A sample from this posterior is then mapped through a neural
network
\begin{equation}
\hat{\boldsymbol{\theta}}_i = g_\phi\!\left(\mathbf{z}_i^{(\theta)}\right),
\end{equation}
yielding the instance-specific solver parameters for the physics model
$\mathcal{F}$.

\paragraph{Discrepancy Latent Encoder.}
Similar to the physics latent encoder, the discrepancy latent encoder maps the context set $\mathcal{C}_i$ to the
parameters of a variational posterior over the discrepancy latent variable
$\mathbf{z}_i^{(\delta)}$:
\begin{equation}
q\!\left(\mathbf{z}_i^{(\delta)} \mid \mathcal{C}_i\right)
= \mathcal{N}\!\left(\boldsymbol{\mu}_\delta(\mathcal{C}_i), \mathbf{\Sigma}_\delta(\mathcal{C}_i)\right),
\end{equation}
where $\boldsymbol{\mu}_\delta(\cdot)$ and $\mathbf{\Sigma}_\delta(\cdot)$ are neural
networks shared across instances. The latent variable $\mathbf{z}_i^{(\delta)}$
captures instance-specific variations in the structured discrepancy, representing
systematic deviations not explained by the physics solver.

A sample from $\mathbf{z}_i^{(\delta)}$, together with the physics latent $\mathbf{z}_i^{(\theta)}$ and a target state $\mathbf{x}_i^{(t)}$, is provided to the discrepancy function
\begin{equation}\label{eq:discrepancy}
\boldsymbol{\delta}_i(\mathbf{x}_{i}^{(t)}; \mathbf{z}_i^{(\theta)}, \mathbf{z}_i^{(\delta)})
= f_\psi\!\left(\mathbf{x}_i^{(t)}, \mathbf{z}_i^{(\theta)}, \mathbf{z}_i^{(\delta)}\right).
\end{equation}
$f_\psi$ is modeled as a neural network and represents systematic discrepancies shared across realizations, while allowing instance-specific corrections that depend on the system state and physical parameters.

\paragraph{Generative Model.}

Given the inferred latent variables $\mathbf{z}_i^{(\theta)}$ and $\mathbf{z}_i^{(\delta)}$,
the decoder defines the conditional distribution of an observation
$\mathbf{y}_i^{(t)}$ at a target input $\mathbf{x}_i^{(t)}$ as
\begin{equation}
\begin{aligned}
p(\mathbf{y}_i^{(t)} \mid \mathbf{x}_i^{(t)}, \mathbf{z}_i^{(\theta)}, \mathbf{z}_i^{(\delta)})
= \\
\mathcal{N}\Big(
\mathcal{F}(\mathbf{x}_i^{(t)}; \hat{\boldsymbol{\theta}}_i)
+
\boldsymbol{\delta}(\mathbf{x}_i^{(t)}; \mathbf{z}_i^{(\delta)}, \boldsymbol{z}_i^{(\theta)}),
\, \Sigma
\Big),
\end{aligned}
\end{equation}
where $\Sigma$ is the observation noise.

The posterior predictive distribution conditioned on the context set is
\begin{equation}
\begin{aligned}
p(\mathbf{y}_i^{(t)} \mid \mathbf{x}_i^{(t)}, \mathcal{C}_i)
=
\int
p(\mathbf{y}_i^{(t)} \mid \mathbf{x}_i^{(t)}, \mathbf{z}_i^{(\theta)}, \mathbf{z}_i^{(\delta)}) \\
\, q(\mathbf{z}_i^{(\theta)}, \mathbf{z}_i^{(\delta)} \mid \mathcal{C}_i)
\, d\mathbf{z}_i^{(\theta)} d\mathbf{z}_i^{(\delta)},
\end{aligned}
\end{equation}

where we adopt a mean-field variational approximation,
\begin{equation}
q(\mathbf{z}_i^{(\theta)}, \mathbf{z}_i^{(\delta)} \mid \mathcal{C}_i)
=
q(\mathbf{z}_i^{(\theta)} \mid \mathcal{C}_i)\,
q(\mathbf{z}_i^{(\delta)} \mid \mathcal{C}_i).
\end{equation}
Note that although the approximate posterior factorizes, interaction between the two latent variables is reintroduced through the discrepancy function network (\autoref{eq:discrepancy}), which models their joint influence on the observations.

\subsection{Training and inference}

We train the model using a two-stage procedure to encourage a meaningful
decomposition between physics and discrepancy components.
The training data consists of two sources: (i) a nominal (or reference) dataset and 
(ii) a real observation dataset. The nominal dataset contains realizations 
generated using the physics solver with known parameters and without unmodeled discrepancies. This provides ground-truth supervision for the physics parameters $\mathbf{\theta}$. 
In contrast, the real observation dataset consists of measurements from the 
true system, where discrepancies between the physics solver and reality may 
be present. The nominal data is used to anchor the physics latent to physically 
meaningful parameters during pretraining, while the real observations are used 
to train the full probabilistic model with discrepancy correction.

\paragraph{Stage 1: Physics Pretraining on Nominal Data.}

In the first stage, we train only the physics pathway using a nominal dataset,
for which ground-truth solver parameters $\boldsymbol{\theta}_i$ are available.
The discrepancy network and decoder are disabled during this stage.

We minimize the supervised parameter loss
\begin{equation}
\mathcal{L}_\theta
=
\sum_{i=1}^{N_{\mathrm{nom}}}
\|\boldsymbol{\theta}_i - \hat{\boldsymbol{\theta}}_i\|_2^2,
\end{equation}
where $N_{\mathrm{nom}}$ denotes the number of realizations in the nominal dataset and $\hat{\boldsymbol{\theta}}_i$ is the
predicted solver parameter.
This stage encourages $\mathbf{z}^{(\theta)}$
to encode physically meaningful structure before introducing a discrepancy model.

\paragraph{Stage 2: Joint Variational Training.}

In the second stage, we activate the discrepancy network and decoder, and train the full model via variational inference. 
This stage leverages both the real observation dataset and the nominal dataset.

We place standard normal priors on both latent variables:

\[
p(\mathbf{z}^{(\theta)}) = \mathcal{N}(\mathbf{0}, \mathbf{I}), \qquad
p(\mathbf{z}^{(\delta)}) = \mathcal{N}(\mathbf{0}, \mathbf{I}).
\]

For each realization $i$ in the real observation dataset, we maximize the evidence lower bound (ELBO):
\begin{equation}
\begin{aligned}
\mathcal{L}^{\mathrm{ELBO}}_{i} =
\mathbb{E}_{q(\mathbf{z}_i^{(\theta)} \mid \mathcal{C}_i)\, q(\mathbf{z}_i^{(\delta)} \mid \mathcal{C}_i)}
\big[
\log p(\mathbf{y}_i^{(t)} \mid \mathbf{x}_i^{(t)}, \mathbf{z}_i^{(\theta)}, \mathbf{z}_i^{(\delta)})
\big] \\
-
\mathrm{KL}\big(q(\mathbf{z}_i^{(\theta)} \mid \mathcal{C}_i)\,\|\,p(\mathbf{z}_i^{(\theta)})\big)
-
\mathrm{KL}\big(q(\mathbf{z}_i^{(\delta)} \mid \mathcal{C}_i)\,\|\,p(\mathbf{z}_i^{(\delta)})\big),
\end{aligned}
\end{equation}

To encourage disentanglement between the physics and discrepancy components, we retain the physics parameter supervision from Stage 1 on the nominal dataset to anchor the physics encoder and mapping network. Simultaneously, we regularize the discrepancy function via 
\begin{equation}
    \mathcal{L}_{\delta} = \sum_{i=1}^{N} \mathcal{R}\bigl(\boldsymbol{\delta}_i(\cdot)\bigr).
\end{equation}
$\mathcal{R}$ is a functional penalty that encodes prior assumptions about the discrepancy structure. This can include standard $L_1$/ $L_2$ penalties or more structured regularizers such as orthogonality or smoothness constraints~\citep{gulrajani2017improved, yoshida2017spectral}. The regularization discourages the discrepancy from absorbing variation that can be explained by the physics encoder, thereby encouraging disentanglement between physics and residual corrections. Based on empirical evidence from ablation studies, we found that $L_2$ regularization provides the best trade-off between predictive performance and disentanglement, and is therefore used in all experiments.
The final objective is
\begin{equation}
\mathcal{L}
=
\sum_{i=1}^N
-\mathcal{L}^{\mathrm{ELBO}}_{i}
+
\lambda_\delta \mathcal{L}_\delta + \lambda_\theta \mathcal{L}_\theta,
\end{equation}
 where $\lambda_\theta$, $\lambda_\delta$ are hyperparameters controlling the relative weight of each term.

\paragraph{Inference.} At test time, given a new realization with sparse context set $\mathcal{C}_i$, 
the encoders infer approximate posterior distributions over $\mathbf{z}_i^{(\theta)}$ 
and $\mathbf{z}_i^{(\delta)}$. Samples from these distributions are mapped to solver 
parameters $\hat{\boldsymbol{\theta}}_i$ and propagated through $\mathcal{F}$, while 
the discrepancy network produces state-dependent corrections. Monte Carlo samples 
of the combined output yield predictive means and variances, providing uncertainty 
estimates over the target observations.
\begin{table*}[t]
\caption{Quantitative comparison for damped spring system. $\dagger$ methods are corrected per realization and not amortized as our method. Best results are shown in bold, and the second-best results are underlined.}
    \begin{adjustbox}{width=0.95\textwidth, center}
    \centering
    \begin{tabular}{lccccccc}
    \toprule
       \multirow{2}{*}{Method} & \multicolumn{3}{c}{Reconstruction} & \multicolumn{3}{c}{Discrepancy ($\delta$) prediction} & \multicolumn{1}{c}{Parameter ($\theta$) prediction} \\ 
        \cmidrule(lr){2-4} \cmidrule(lr){5-7} 
        & MAE ($\downarrow$) & LL ($\uparrow$) & ECE ($\downarrow$) & MAE ($\downarrow$) & LL ($\uparrow$) & ECE ($\downarrow$) & MAE ($\downarrow$) \\
    \midrule
       KOH-GP$^\dagger$ &  0.017 $\pm$ 0.009   & 0.738 $\pm$ 0.086 & 0.413 $\pm$ 0.010  & 0.760 $\pm$ 0.118    & 0.338 $\pm$ 0.049  & 0.411 $\pm$ 0.016 & 2.957 $\pm$ 1.487  \\
       BCPI$^\dagger$ & 0.033 $\pm$  0.016  &   1.079 $\pm$ 0.267   & 0.206 $\pm$ 0.035   & 0.621 $\pm$ 0.272 & 0.916 $\pm$ 0.122& 0.475 $\pm$ 0.051& 1.395 $\pm$ 0.512  \\ 
       Correction-PINN$^\dagger$ &  0.094 $\pm$ 0.068   & -0.718 $\pm$ 0.203   & 0.471 $\pm$ 0.021   & 1.546 $\pm$ 0.675   & -0.771 $\pm$ 0.278 & 0.420 $\pm$ 0.043   &  0.654 $\pm$ 0.375\\ \midrule
       APIC-CNP  & \textbf{0.013$\pm$0.003} & \textbf{1.133$\pm$0.004} & 0.247$\pm$ 0.001 & \textbf{0.009$\pm$0.006} & \underline{1.130$\pm$0.006} & 0.298$\pm$0.032 & \underline{0.097$\pm$0.020}    \\
       APIC-LNP & \underline{0.015$\pm$ 0.006} & \underline{1.121$\pm$0.017} & \textbf{0.128$\pm$0.027}  & \underline{0.011$\pm$0.006} & \textbf{1.136$\pm$0.009} & \textbf{0.168$\pm$0.008} & \textbf{0.081$\pm$0.028} \\ 
       APIC-ANP & 0.023$\pm$0.002 & 0.932$\pm$0.027 & \underline{0.147$\pm$0.021} & 0.037$\pm$0.004 & 0.663$\pm$0.051 & \underline{0.241$\pm$0.063} & 0.104$\pm$0.060   \\ 
    \bottomrule
    \end{tabular}
    \label{tab:damped_spring_id}
    \end{adjustbox}
\end{table*}

\begin{figure*}
    \centering
    \includegraphics[width=\linewidth]{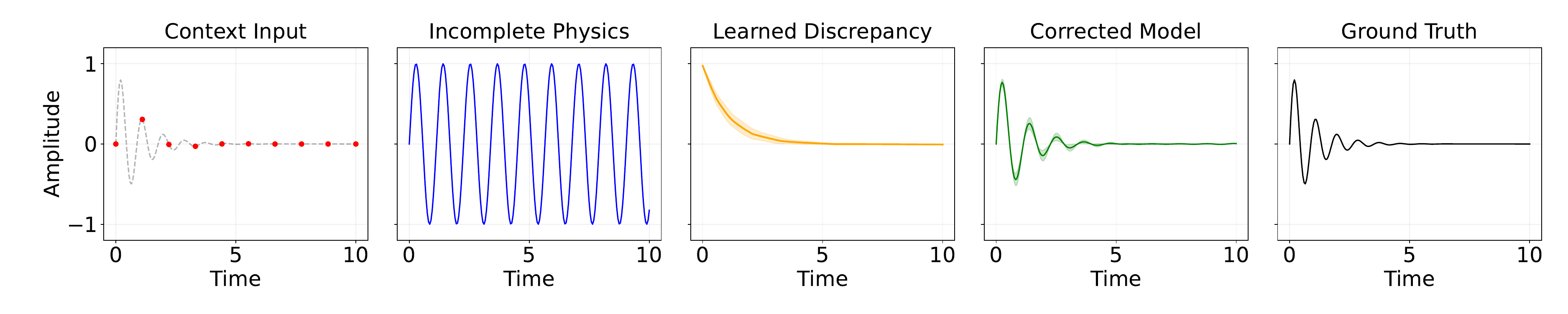}
    
    \caption{Qualitative correction pipeline for the damped spring system. From left to right: sparse context observations; prediction from the misspecified nominal simulator; learned discrepancy; corrected prediction obtained by combining simulator and discrepancy; and the ground-truth solution. APIC identifies and compensates for the structured model bias.}
    \label{fig:damped_spring}
\end{figure*}

\section{Experiments}

\paragraph{Experimental Setup.} 
We consider three benchmark systems: a spring oscillator, a Lotka-Volterra ODE, and an advection-diffusion PDE. 
For each benchmark, the learner has access to a nominal physics simulator, while observations are generated from a true process that includes a structured, state-dependent discrepancy. 
The discrepancy is systematic rather than random noise, and it depends on task-specific quantities such as physical parameters or initial conditions. 
This setup reflects realistic scientific scenarios in which simulators capture known physics but omit structured, unmodeled effects. All models are implemented in PyTorch. 

\paragraph{Baselines.}
We compare against three physics model calibration baselines.
The first baseline is the classical KOH framework \cite{KOH2001}, where the discrepancy is modeled as a Gaussian Process \cite{Rasmussen2006Gaussian}. We use a zero-mean RBF kernel and estimate the physical parameters ($\theta$) together with the GP hyperparameters, via marginal likelihood maximization for each realization. Predictive uncertainty is obtained from the GP posterior. Bayesian Calibration with Physics-Informed Priors (BCPI; \cite{BCPI}) instantiates a KOH-style model but incorporates physics-informed priors within the GP to improve identifiability and extrapolation. Correction-PINN \cite{ZOU2024112918} is a PINN-based method in which a secondary correction network augments the governing equations or residual terms, and predictive uncertainty is approximated using deep ensembles of independently trained models.

To evaluate the influence of the NP backbone, we instantiate APIC with three architectures: CNP, LNP, and ANP. In all variants, the physics-informed structure, dual-latent formulation, and discrepancy modeling remain identical, and only the context aggregation mechanism differs.
\textbf{APIC-CNP} uses a global mean aggregation and a deterministic encoding~\citep{garnelo2018conditional}, yielding a fixed task representation.   
\textbf{APIC-LNP} and \textbf{APIC-ANP} both employ stochastic task-level embeddings as described in \autoref{sec:method}. The difference lies in the aggregation mechanism: APIC-LNP uses global mean aggregation in the encoder~\citep{garnelo2018neural}, whereas APIC-ANP employs attention-based aggregation to obtain target-dependent context weighting~\citep{kim2019attentive}. Models are trained using the Adam optimizer with learning rate $5\times10^{-4}$. The values of $\lambda_\theta$ and $\lambda_\delta$ are set to 1. Details on the model architecture and training hyperparameters are provided in \autoref{app:implementation}.

\paragraph{Evaluation Metrics.}
We evaluate performance along three axes: predictive reconstruction, discrepancy identification, and parameter recovery.
For predictive reconstruction, we report the Mean Absolute Error (MAE) to measure pointwise reconstruction error and the Predictive Log-Likelihood (LL) to measure the probabilistic accuracy of the predictive distributions. Uncertainty calibration is assessed using the Expected Calibration Error (ECE), computed as the weighted discrepancy between nominal and empirical coverage across multiple confidence levels~\cite{kuleshov2018accurate}. 
For discrepancy identification, we compare the inferred discrepancy function to the ground-truth discrepancy and report MAE and LL evaluated over target inputs. Calibration of discrepancy uncertainty is likewise quantified using ECE.
Finally, parameter recovery is evaluated using MAE between the inferred parameters and the ground-truth parameters.

\begin{figure*}[t]
    \centering
    \includegraphics[width=0.9\linewidth]{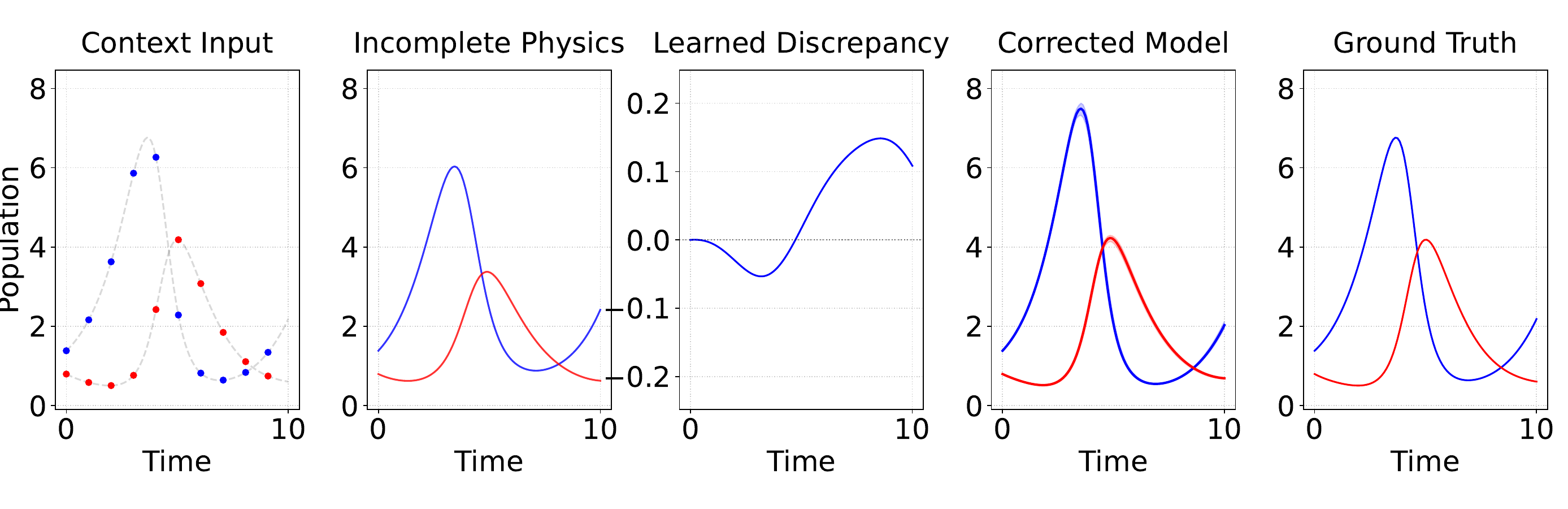}
    \caption{Qualitative correction pipeline for the Lotka-Volterra ODE System. From left to right: sparse context observations; prediction from the misspecified nominal simulator; learned discrepancy; corrected prediction obtained by combining simulator and discrepancy; and the ground-truth solution. The blue plot refers to the \textcolor{blue}{prey} and the red plot is the \textcolor{red}{predator}. }
    \label{fig:lv_reconstruction_full}
    
\end{figure*}

\begin{table*}[t]
\caption{Quantitative results for Lotka-Volterra ODE System. $\dagger$ methods are corrected per realization and not amortized as our method. Best results are shown in bold, and the second-best results are underlined.}
    \begin{adjustbox}{width=0.95\textwidth, center}
    \centering
    \begin{tabular}{lccccccc}
    \toprule
       \multirow{2}{*}{Method} & \multicolumn{3}{c}{Reconstruction} & \multicolumn{3}{c}{Discrepancy ($\delta$) prediction} & \multicolumn{1}{c}{Parameter ($\theta$) prediction} \\ 
        \cmidrule(lr){2-4} \cmidrule(lr){5-7} 
        & MAE ($\downarrow$) & LL ($\uparrow$) & ECE ($\downarrow$) & MAE ($\downarrow$) & LL ($\uparrow$) & ECE ($\downarrow$) & MAE ($\downarrow$)\\
    \midrule
       KOH-GP$^\dagger$  & 0.197 $\pm$ 0.066 & 0.163 $\pm$ 0.026 &0.439 $\pm$ 0.008 & 0.197 $\pm$ 0.115   & 0.668 $\pm$ 0.438  & 0.347 $\pm$ 0.045&  0.163 $\pm$ 0.120 \\
       BCPI$^\dagger$   & 0.484 $\pm$ 0.028  & -4.235 $\pm$ 0.917 & 0.436 $\pm$ 0.056  & 0.284 $\pm$ 0.101   & -4.762 $\pm$ 0.265 & 0.492 $\pm$ 0.008 & 0.283 $\pm$ 0.118 \\
       Correction-PINN$^\dagger$ & 0.088 $\pm$ 0.045  & -4.834 $\pm$ 1.305 & 0.290 $\pm$ 0.076 & 0.301 $\pm$ 0.076   & -3.030 $\pm$ 0.294  & 0.488 $\pm$ 0.020 &  0.147 $\pm$ 0.037 \\ \midrule
       APIC-CNP  & 0.544$\pm$0.086 & 0.680$\pm$0.329 & \textbf{0.143$\pm$0.005} & \underline{0.048$\pm$0.007}   & 0.085$\pm$0.016  & \textbf{0.206$\pm$0.007} & \textbf{0.049$\pm$0.006}  \\
       APIC-LNP & \textbf{0.036$\pm$0.003} & \textbf{1.454$\pm$0.177} & 0.251$\pm$0.009 & \textbf{0.030$\pm$0.007}   &  \underline{1.268$\pm$0.153} & 0.279$\pm$0.042 & \underline{0.059$\pm$0.005}  \\
       APIC-ANP & \underline{0.061$\pm$0.005} & \underline{1.369$\pm$0.164} & \underline{0.248$\pm$0.003} & \underline{0.048$\pm$0.007}  & \textbf{1.738$\pm$0.071}  & \underline{0.273$\pm$0.016} & 0.102$\pm$0.016  \\ 
    \bottomrule
    \end{tabular}
    \label{tab:lotka_volterra_id}
    \end{adjustbox}
\end{table*}

\subsection{Damped spring system}

\paragraph{Dataset. }The nominal simulator $\mathcal{F}$ corresponds to an undamped harmonic oscillator, $u''(t) + (\omega )^2 u(t) = 0$. The initial conditions are chosen such that the analytical solution is $\mathcal{F}(t; \omega ) = \sin(\omega t)$, where the frequency $\omega$  is sampled independently per task from $\mathcal{U}(3,8)$.
The true data-generating process introduces parameter-dependent exponential damping, $\mathcal{G}(t; \omega ) 
= \sin(\omega  t)\exp(-\lambda \omega  t),
\lambda = 0.15$. This introduces a structured discrepancy
$\delta (t) 
= \mathcal{G}(t; \omega ) - \mathcal{F}(t; \omega )
= \sin(\omega  t)\big(\exp(-\lambda \omega  t) - 1\big)$.
Trajectories are evaluated on $t \in [0,4]$ using $T=100$ uniformly spaced points.
For each task, observations are partitioned into a context set of randomly sampled pairs and a target set comprising the remaining time points. During training, the context size is uniformly sampled from 15 to 100, while at test time, we randomly sample 30 context points.
Results are averaged over 200 independent function realizations.

\paragraph{Results.} Quantitative results are reported in \autoref{tab:damped_spring_id}. 
All APIC variants outperform classical calibration methods across reconstruction, discrepancy prediction, and parameter recovery. We observe that APIC achieves a significantly lower discrepancy and parameter MAE while maintaining a higher predictive log-likelihood. 
Moreover, classical calibration methods are non-amortized; predictions are computed per sample, whereas APIC performs amortized calibration. 
This indicates that disentangling physical parameters from structured discrepancy mitigates bias under model misspecification.
The stochastic variants (APIC-LNP and APIC-ANP) also exhibit improved uncertainty calibration, as demonstrated by lower ECE scores. \autoref{fig:damped_spring} provides a qualitative analysis of a sample test trajectory using APIC-LNP. The first panel shows the sparse spatio-temporal context observations. The second panel presents the prediction obtained from the misspecified physics model, which fails to capture the systematic damping term. 
The third panel displays the learned discrepancy field, which accurately identifies the structured bias between the nominal simulator and the true dynamics. When this correction is combined with the simulator output (fourth panel), the resulting prediction closely matches the ground-truth field shown in the final panel. 
These results demonstrate that APIC successfully isolates and learns structured model error, enabling accurate recovery of the full spatio-temporal dynamics despite misspecified physics and sparse observations.

In terms of computational time, during inference, APIC-LNP requires only 0.42 ms per realization at test time, whereas BCPI requires 12.6 s, KOH-GP requires 33.6 s, and Correction-PINN requires 150 s (with 20 ensembles) per realization, which represents a reduction of approximately 4-5 orders of magnitude.

Ablation analyses on the impact of context point size, observation noise, regularization, $\lambda_\theta$ and $\lambda_\delta$ are provided in Appendix~\ref{app:ablation}.

\subsection{Lotka--Volterra ODE system}

\paragraph{Dataset.}
The nominal simulator $\mathcal{F}$ corresponds to the Lotka--Volterra system with state $u(t)=(x(t),y(t))$. $x(t)$ denotes the prey population and $y(t)$ denotes the predator population.
\[
\dot{x}=\alpha  x - \beta  xy, \qquad 
\dot{y}=\eta  xy - \gamma  y.
\]
The task-specific parameters $\boldsymbol{\theta} =(\alpha ,\beta ,\eta ,\gamma )$ are independently sampled from $\mathcal{U}(0.5,1.5)$ along with the initial condition $u_0 \sim \mathcal{U}(0.7,1.8)$. 
The true dynamics $\mathcal{G}$ introduce a structured additive forcing in the prey channel,
\[
\dot{x}=\alpha  x - \beta  xy + \delta_x (t), \qquad
\dot{y}=\eta  xy - \gamma  y,
\]
where $\delta_x (t)=\Phi(t)^\top w $ is parameterized by an RBF time basis $\Phi(t)$ with $\Phi(0)=0$. The coefficients $w $ are generated via a fixed systemic map from task features $[\log(\boldsymbol{\theta} ),u_0]$, yielding structured, task-dependent discrepancy with $\delta_x (0)=0$. 
Both systems are integrated using RK4 on $t\in[0,10]$ with $T=100$ time steps.
At training, the context size is uniformly sampled from 20 to 100 and fixed to 30 at test time. Results are averaged over 200 independently sampled tasks.

\paragraph{Results.} \autoref{tab:lotka_volterra_id} provides quantitative results. APIC-LNP achieves the best overall performance, with the lowest reconstruction MAE  and highest log-likelihood. At the same time, it accurately captures the discrepancy and recovers parameters. APIC-ANP performs competitively with APIC-LNP, whereas APIC-CNP exhibits a lower performance in terms of reconstruction and discrepancy MAE and LL, possibly due to its deterministic aggregation. Classical per-instance methods show reasonable reconstruction but higher discrepancy and parameter errors. Qualitative reconstructions for the correction pipeline for predator and prey are provided in \autoref{fig:lv_reconstruction_full} using APIC-LNP. APIC produces predictions closely aligned with the ground truth. Additional qualitative results for uncertainty and discrepancy estimation are provided in Appendix~\ref{app:lv_qualitative}.

\subsection{Advection-diffusion PDE}

\begin{figure*}[t]
    \centering
    \includegraphics[width=0.95\linewidth]{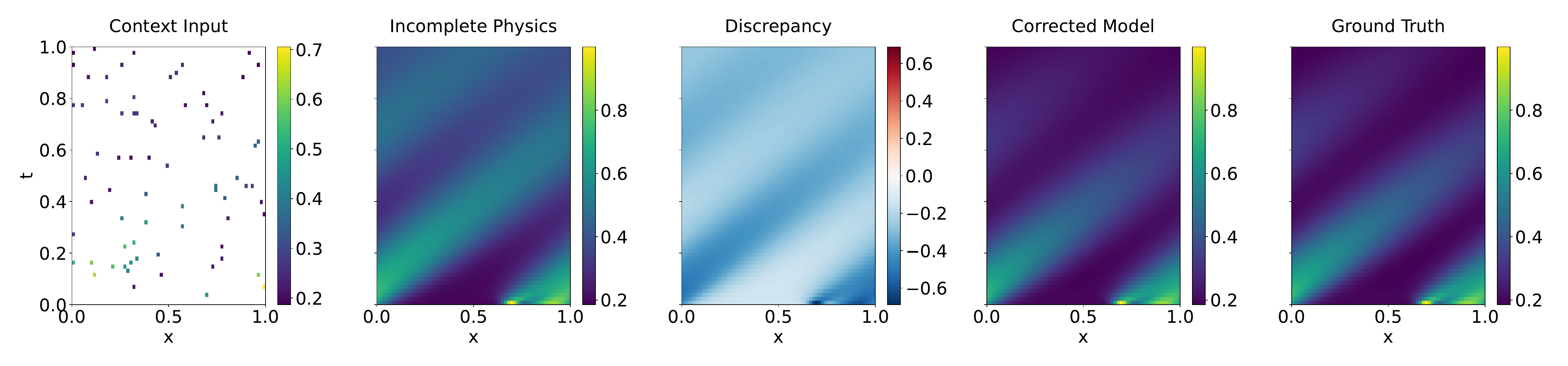}
    \caption{Qualitative correction pipeline for the advection–diffusion system. From left to right: sparse context observations; prediction from the misspecified nominal simulator; learned discrepancy field; corrected prediction obtained by combining simulator and discrepancy; and the ground-truth solution. APIC identifies and compensates for the structured model bias, recovering the true spatio-temporal dynamics.}

    \label{fig:adv_diff_qualitative_full}
\end{figure*}

\begin{table*}[t]
\caption{Quantitative results for advection-diffusion system. $\dagger$ methods are corrected per realization and not amortized as our method. Best results are shown in bold, and the second-best results are underlined.}
    \begin{adjustbox}{width=0.95\textwidth, center}
    \centering
    \begin{tabular}{lcccccccc}
    \toprule
       \multirow{2}{*}{Method} & \multicolumn{3}{c}{Reconstruction} & \multicolumn{3}{c}{Discrepancy ($\delta$) prediction} & \multicolumn{2}{c}{Parameter ($\theta$) prediction MAE} \\ 
        \cmidrule(lr){2-4} \cmidrule(lr){5-7} \cmidrule{8-9}  
        & MAE ($\downarrow$) & LL ($\uparrow$) & ECE ($\downarrow$) & MAE ($\downarrow$) & LL ($\uparrow$) & ECE ($\downarrow$) & $\kappa$($\downarrow$) & $v$ ($\downarrow$)\\
    \midrule
       KOH-GP$^\dagger$ &0.128 $\pm$ 0.047  & 0.276 $\pm$ 0.179 & 0.420 $\pm$ 0.017& 0.214 $\pm$ 0.062   & 0.628 $\pm$ 0.083  & 0.382 $\pm$ 0.022& 0.865 $\pm$ 0.187 & 0.759 $\pm$ 0.241  \\
       BCPI$^\dagger$   & 0.031 $\pm$ 0.019  & 0.933 $\pm$ 0.212  & 0.896 $\pm$ 0.357 & 0.362 $\pm$ 0.176  & -1.336 $\pm$ 1.059  & 0.934 $\pm$ 0.395 & 0.020 $\pm$ 0.019& 0.390 $\pm$ 0.294  \\
       Correction-PINN$^\dagger$ & \underline{0.021 $\pm$ 0.003} & -1.552 $\pm $ 0.523 & 0.136 $\pm$ 0.072&  0.070 $\pm$ 0.018  & -3.259 $\pm$ 0.655  &  0.268 $\pm$ 0.100& 0.021 $\pm$ 0.020 & \underline{0.121 $\pm$ 0.086}  \\ \midrule
       APIC-CNP  & 0.054$\pm$0.011 & 1.177$\pm$0.186 & \textbf{0.041$\pm$0.008} & \underline{0.025$\pm$0.014}   & 1.475$\pm$0.178  & \textbf{0.170$\pm$0.060} &  \underline{0.019$\pm$0.001} & 0.371$\pm$0.084 \\
       APIC-LNP & \textbf{0.010$\pm$0.002} & \textbf{2.713$\pm$0.224} & 0.064$\pm$0.004 &  \textbf{0.007$\pm$0.006}  & \textbf{2.925$\pm$0.412}  & 0.209$\pm$0.045 &  \textbf{0.004$\pm$0.000} & \textbf{0.059$\pm$0.008} \\
       APIC-ANP & 0.053$\pm$0.012 & \underline{1.183$\pm$0.191} & \underline{0.043$\pm$0.010} &  0.026$\pm$0.014  & \underline{1.502$\pm$0.167}  & \underline{0.173$\pm$0.049} & \underline{0.019$\pm$0.001} & 0.361$\pm$0.091  \\ 

    \bottomrule
    \end{tabular}
    \label{tab:adv_diff_id}
    \end{adjustbox}
\end{table*}

\paragraph{Dataset.} We consider a periodic one-dimensional field $u(t,x)$ on $x \in [0,1)$ governed by the nominal advection–diffusion equation
$$u_t + v  u_x = \kappa  u_{xx}, 
\qquad 
\boldsymbol{\theta}  = (v , \kappa ),$$
where $v  \sim \mathcal{U}(0.3,2.0)$ and $\kappa  \sim \mathcal{U}(0.002,0.08)$ are sampled independently per task, and the initial condition $u(0,x)=u_0(x)$ is a smooth random field. 

The true data-generating process includes an additional parameter-dependent reaction (sink) term,

$$u_t + v  u_x = \kappa  u_{xx} - c (\boldsymbol{\theta} ) u, $$
where the reaction rate follows a systematic law 
$c (\boldsymbol{\theta} )=\mathrm{softplus}\!\big(a^\top\log(\boldsymbol{\theta} )+b\big),$
clipped to a maximum value. This induces a structured discrepancy 
$\delta (t,x)=\mathcal{G}(t,x;\boldsymbol{\theta} )-\mathcal{F}(t,x;\boldsymbol{\theta} ).$

Both systems are solved using an exact Fourier spectral method under periodic boundary conditions. Solutions are evaluated on a uniform grid with $N_t=64$ time steps and $N_x=64$ spatial points. For each task, context observations are randomly subsampled from the full spatio-temporal grid, and performance is evaluated on the remaining target points.

\paragraph{Results.}
\autoref{tab:adv_diff_id} reports quantitative results. APIC-LNP achieves the best overall performance. 
Parameter recovery further highlights the advantage of amortized inference. APIC-LNP significantly outperforms classical per-instance approaches, particularly in estimating the advection coefficient $v$, where errors are reduced by an order of magnitude relative to KOH-GP. 
APIC-CNP and APIC-ANP yield comparable performance. In contrast, per-instance calibration methods exhibit larger parameter and calibration errors, especially in discrepancy prediction.
\autoref{fig:adv_diff_qualitative_full} illustrates the full correction pipeline for the advection–diffusion system. We see that the resulting prediction after accounting for the model discrepancy closely matches the ground-truth field shown in the final panel. Additional qualitative analyses are provided in Appendix~\ref{app:lv_qualitative}.

\section{Conclusion}

We introduced APIC, a population-level extension of the calibration principle in \cite{KOH2001} that replaces per-realization Bayesian calibration with amortized inference across related systems. Instead of solving a new inverse problem for each realization, APIC leverages a dual-latent architecture Neural Process \cite{garnelo2018neural} backbone to map sparse context observations to predictive distributions while separating instance-specific physics parameters from structural discrepancy. 
Experimental results show that APIC improves reconstruction and discrepancy identification while preserving accurate parameter recovery and calibrated uncertainty in low-data regimes. 
These findings support amortized calibration as an effective organizing principle for hybrid modeling, reducing the need for repeated per-instance optimization while enabling adaptation of imperfect models to observed data.
APIC assumes differentiable simulator models, sufficient simulation data for meta-training, and appropriate choices of discrepancy parameterization, which may limit applicability in black-box, simulation-scarce, or structurally misspecified settings. Addressing these constraints provides a natural direction for future work. We discuss this further in Appendix~\ref{sec:ap_lim}.

\FloatBarrier

\bibliography{uai2026-template}

\clearpage
\newpage

\onecolumn

\title{APIC: Amortized Physics-Informed Calibration using Neural Processes\\ (Supplementary Material)}
\maketitle

\appendix

\section{Implementation Details}\label{app:implementation}


\paragraph{Damped Spring System.}
For the damped spring system, we employ two independent multilayer perceptron (MLP) latent encoders for physics and discrepancy inference. Each encoder first flattens the context trajectory and processes it through a two-layer MLP with hidden dimensions $128 \rightarrow 64$, producing a hidden embedding of dimension $64$. Linear heads output the mean and log-variance of 1-dimensional Gaussian latent variables.
The physical frequency $\omega$ is inferred from the physics latent via a two-layer MLP with hidden dimension 128. The discrepancy function is modeled multiplicatively. The predictive distribution is Gaussian with learned mean and variance. Observation noise in the decoder is modeled with a Gaussian distribution with diagonal covariance $\Sigma$.
Training proceeds in two stages. In the first stage (3{,}000 steps), the model is trained to recover physical parameters from trajectories without discrepancy correction to anchor parameter identification. In the second stage (4{,}000 steps), full variational training is performed, jointly optimizing parameter inference and discrepancy modeling. Both stages use a batch size of 64.

\paragraph{Lotka--Volterra ODE System.}
We employ a stochastic MLP encoder that maps observed predator--prey trajectories to a Gaussian latent variable of dimension $64$. The encoder first flattens the input trajectory and processes it through a three-layer MLP with hidden dimensions $512 \rightarrow 256 \rightarrow 256$, producing a hidden embedding of dimension $256$. Linear heads produce the mean and log-variance of the latent distribution.
Physical parameters $\theta=(\alpha,\beta,\delta,\gamma)$ are predicted from the latent via a linear decoder in log-space with bounded sigmoid scaling. The discrepancy head consists of a two-layer MLP (hidden size 128) with Softplus activation. 
Corrected dynamics are integrated using a differentiable RK4 solver. Observation noise in the decoder is modeled with a Gaussian distribution with diagonal covariance $\Sigma$. The first stage is trained for 3000 steps, and the second stage for 4000 steps. The batch size is 64 for both stages.

\paragraph{Advection-Diffusion PDE.}
For the advection-diffusion benchmark, the latent encoders employ a three-layer MLP with hidden dimensions $512 \rightarrow 256 \rightarrow 256$, producing a hidden embedding of dimension $d_h=256$. A linear layer maps this embedding to produce the mean and log-variance of a Gaussian latent distribution with dimension $64$.
Physical parameters $(v,\kappa)$ are inferred from the latent variable via a linear head, constrained to valid ranges through a sigmoid transformation in log-space. The discrepancy is parameterized through a three-layer MLP with hidden dimension 256 and outputs the mean and variance. Observation noise in the decoder is modeled with a Gaussian distribution with diagonal covariance $\Sigma$. The first stage is trained for 4000 steps, and the second stage for 10000 steps. The batch size is 64 for both stages.

\section{Additional results}
\subsection{Qualitative results}\label{app:lv_qualitative}

\subsubsection{Lotka--Volterra ODE System}
\begin{figure}[h]
    \centering
    \includegraphics[width=0.6\linewidth]{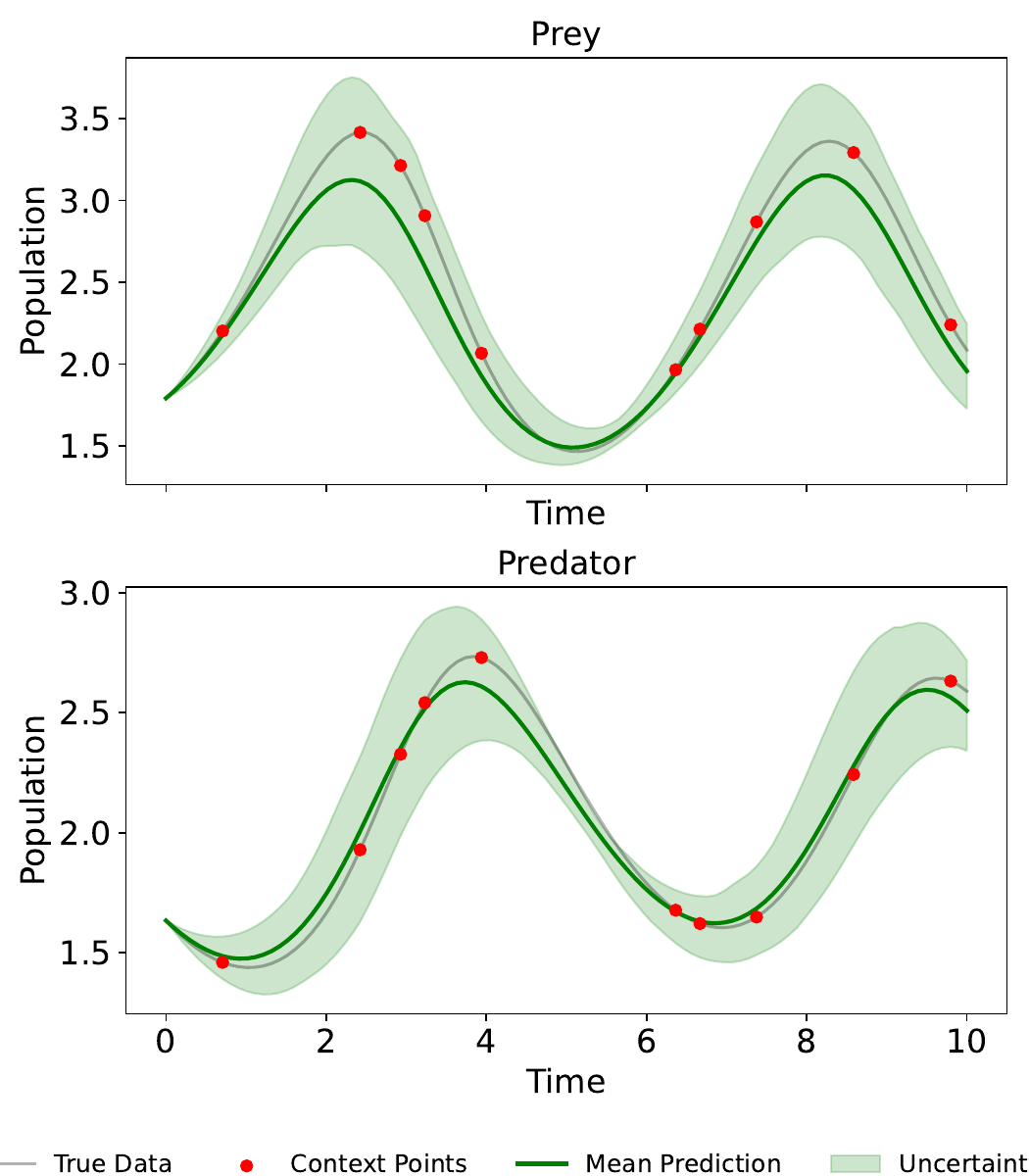}
    \caption{Qualitative reconstruction results for the Lotka--Volterra system. Ground-truth predator and prey populations are shown alongside APIC’s posterior predictive mean and $\pm 2$ standard deviations. The model accurately captures the oscillatory dynamics, and the uncertainty bands reliably cover the true trajectories under sparse observations.}
    \label{fig:lv_reconstruction}
\end{figure}

\begin{figure}[h]
    \centering
    \includegraphics[width=0.7\linewidth]{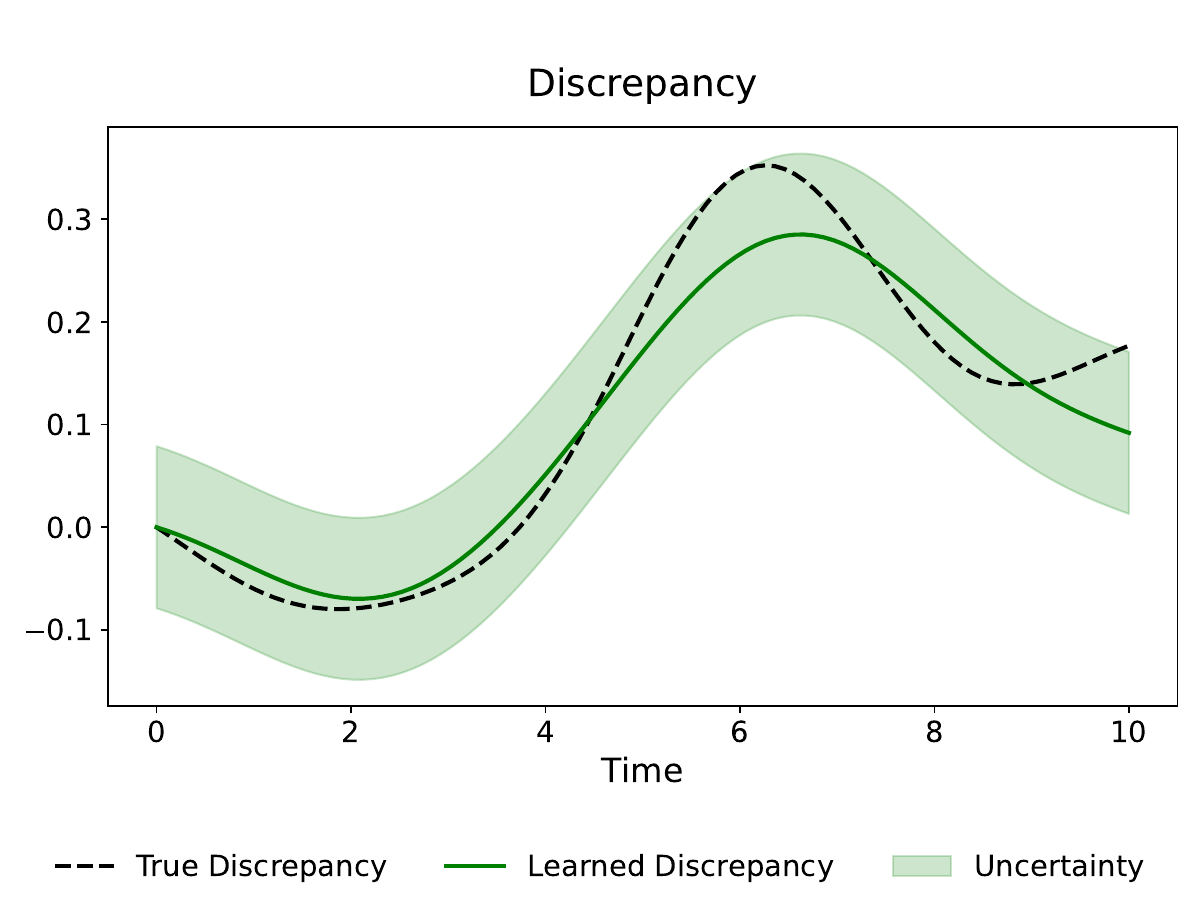}
    \caption{Qualitative discrepancy prediction for the Lotka--Volterra system. The learned discrepancy for predator and prey populations is shown alongside the ground-truth correction and predictive uncertainty ($\pm 2$ standard deviations).}
    \label{fig:lv_discrepancy}
\end{figure}

\autoref{fig:lv_reconstruction} shows the reconstructed prey and predator populations using APIC-LNP, along with the predicted uncertainty and the ground truth. We observe that the model accurately captures the oscillatory dynamics of both populations, and the uncertainty covers the observed data points.
We further examine the learned discrepancy functions for the Lotka–Volterra system. This is illustrated in \autoref{fig:lv_discrepancy}. We observe that APIC can capture the structured deviation between the nominal simulator and the true dynamics. The learned discrepancy closely follows the ground-truth correction across time. Uncertainty bands appropriately widen in regions with limited context observations, while remaining sharp where data is informative. 

\subsubsection{Advection-Diffusion PDE}

\begin{figure}[h]
    \centering
    \includegraphics[width=0.95\linewidth]{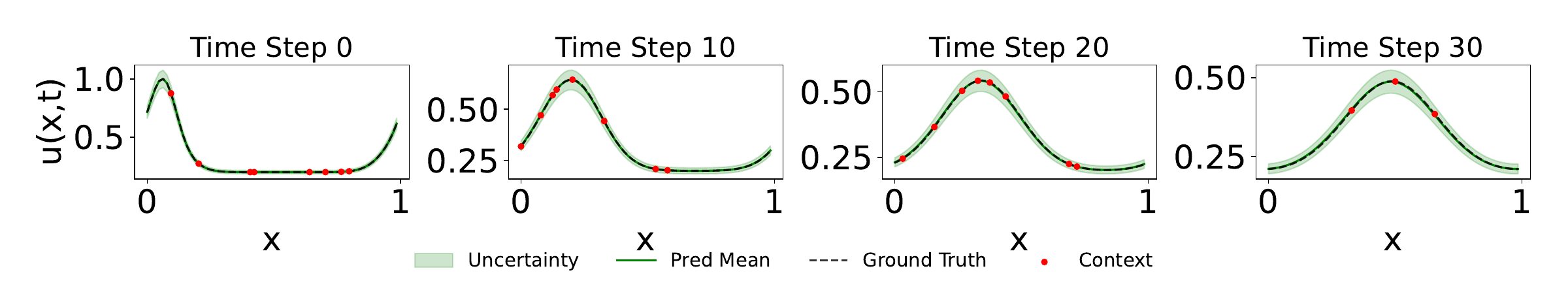}
    \caption{Qualitative reconstruction for the advection–diffusion system using APIC-LNP. Each panel shows the spatial field $u(x, t)$ at different time steps. The predictive mean closely follows the ground truth, while uncertainty bands ($\pm2$ standard deviations) cover the true reconstructions.}
    \label{fig:adv_diff_qualitative_rec}
\end{figure}

\autoref{fig:adv_diff_qualitative_rec} shows qualitative spatial reconstructions at four representative time snapshots using APIC-LNP. It accurately captures the transported and diffused profiles across time, closely matching the ground truth while exhibiting well-calibrated uncertainty. 

\begin{figure}[h]
    \centering
    \includegraphics[width=0.6\linewidth]{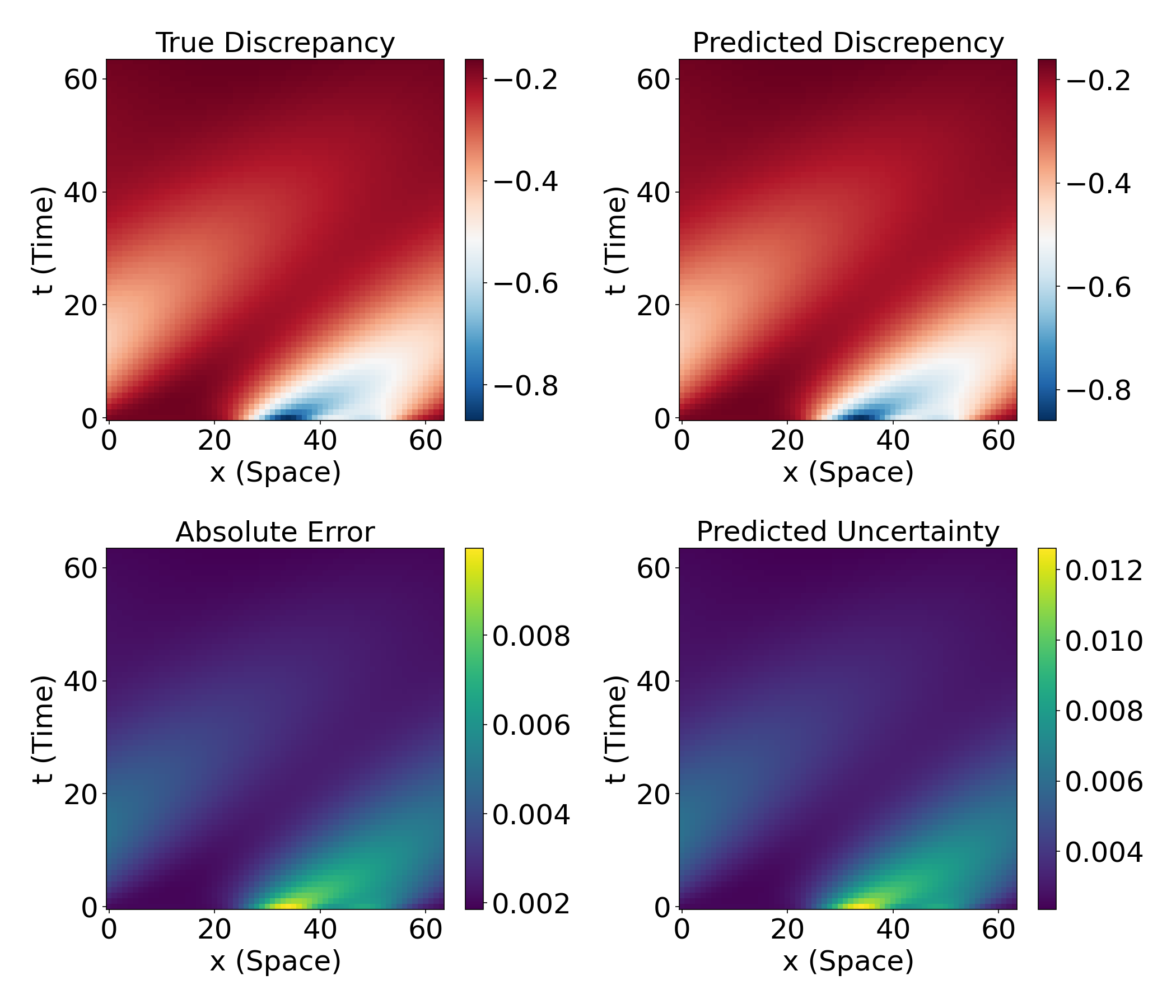}
    \caption{Discrepancy analysis for the advection–diffusion system. The plot shows the predicted discrepancy, the ground-truth discrepancy, the absolute error, and the predictive uncertainty. Regions with higher prediction error correspond to increased uncertainty, indicating calibrated confidence in the learned correction.}
    \label{fig:adv_diff_disc}
\end{figure}

\autoref{fig:adv_diff_disc} further analyzes the learned discrepancy field. We compare the predicted discrepancy with the ground-truth correction and visualize the absolute error between them. The model accurately recovers the dominant spatio-temporal structure of the discrepancy. 
Importantly, regions exhibiting higher absolute error coincide with increased predictive uncertainty, indicating that the model appropriately expresses uncertainty where the discrepancy is more difficult to infer. This alignment between error and uncertainty suggests that the learned posterior provides meaningful and well-calibrated confidence estimates.

\subsection{Ablation Study}\label{app:ablation}
In this study, we assess how the different components of APIC impact its performance. All experiments are performed on the damped-spring system using APIC-LNP.

\paragraph{Effect of Context Set Size.}

\begin{figure}
    \centering
    \includegraphics[width=0.7\linewidth]{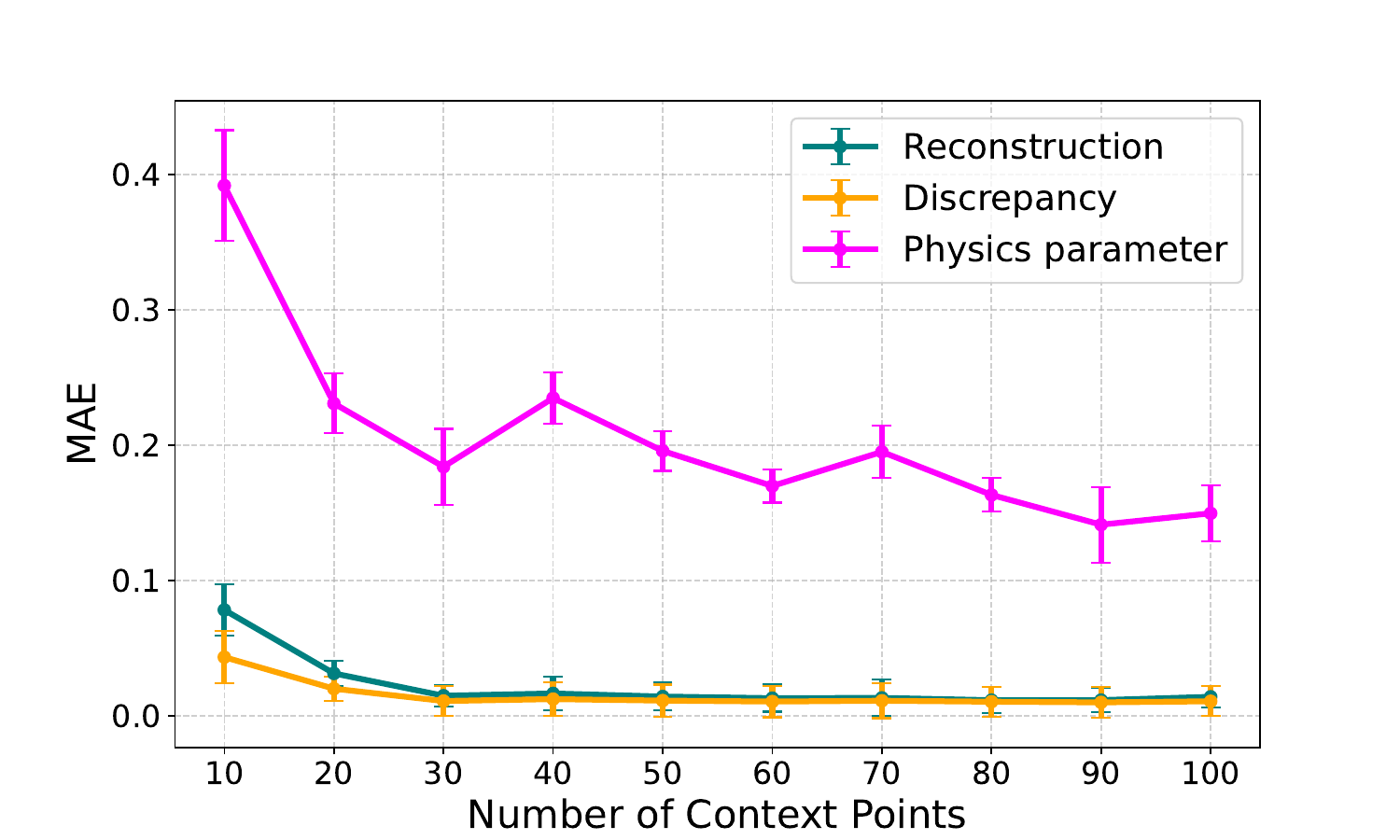}
    \caption{Effect of context set size on predictive performance for the damped spring system. Reconstruction, discrepancy, and parameter MAE are shown as the number of context points increases from 10 to 100. Errors decrease sharply in the low-data regime and stabilize beyond approximately 30 context points.}
    \label{fig:context_ablation}
\end{figure}

We first investigate how the number of context observations influences predictive performance on the damped spring system. The number of context points is varied from 10 to 100, and we report MAE for reconstruction, discrepancy prediction, and physics parameter estimation.

As shown in \autoref{fig:context_ablation}, increasing the number of context points significantly improves performance in the low-data regime. In particular, all three error metrics exhibit a sharp decrease as the context size increases up to approximately 30 points. Beyond this threshold, improvements gradually saturate, and the MAE stabilizes. This suggests that APIC efficiently extracts task-level information from relatively few observations, and that around 30 context points are sufficient to reliably identify both the underlying dynamics and the discrepancy structure for this system.

\begin{figure}
    \centering
    \includegraphics[width=0.7\linewidth]{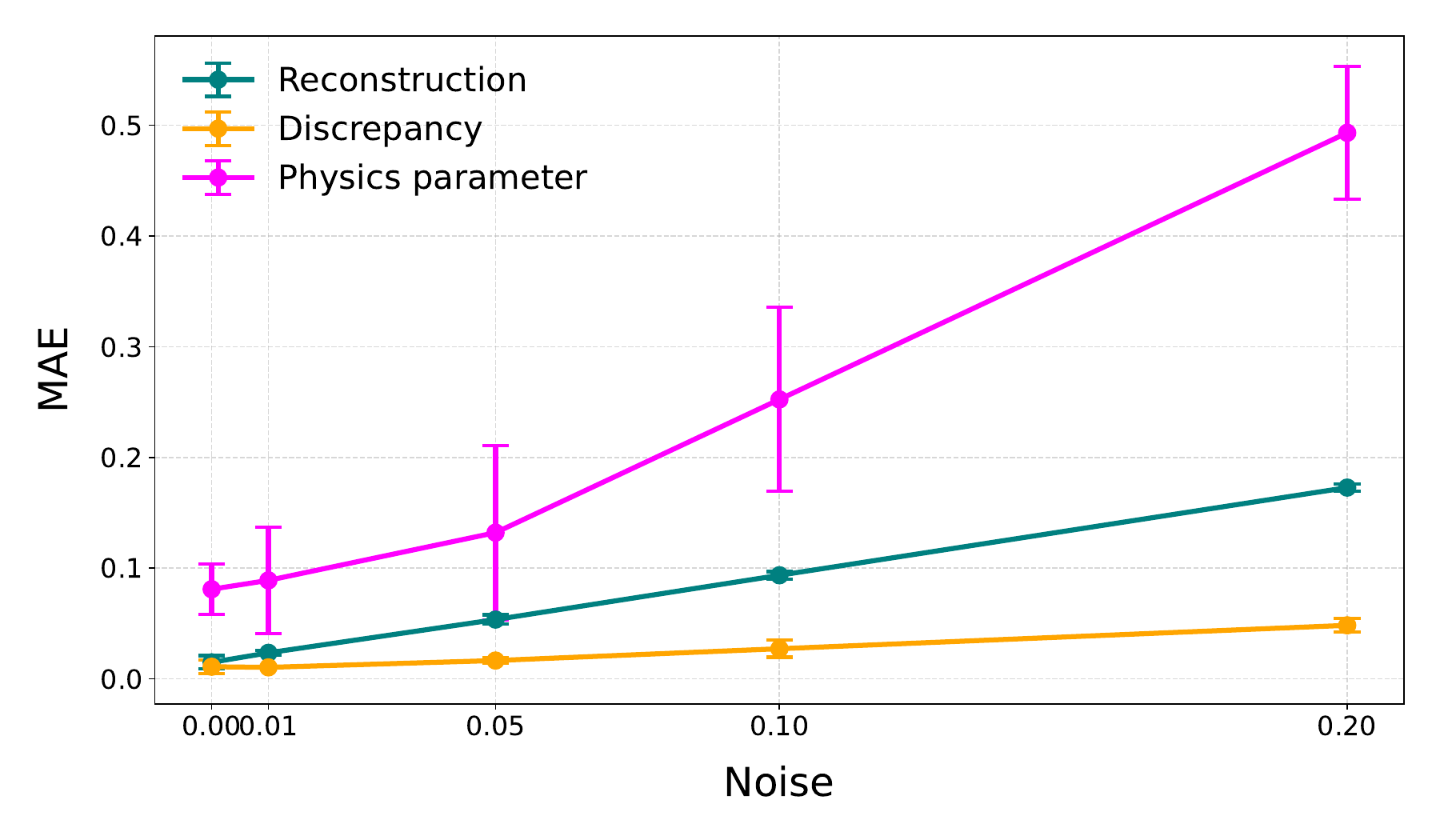}
    \caption{Effect of observation noise on predictive performance for the damped spring system. Reconstruction, discrepancy, and parameter MAE are reported as noise magnitude increases. Errors grow gradually with noise, indicating stable and robust inference under moderate perturbations.}
    \label{fig:noise_ablation}
\end{figure}

\paragraph{Effect of Observation Noise.}
We further evaluate robustness to observation noise by varying the noise level added to the damped spring observations. As shown in \autoref{fig:noise_ablation}, reconstruction, discrepancy, and parameter errors increase progressively with noise magnitude. 

While performance degrades as expected, the increase is gradual rather than abrupt, indicating that APIC maintains stable inference under moderate noise levels. In particular, parameter estimation remains relatively robust in the low-to-medium noise regime, suggesting that the latent-variable formulation effectively separates structural dynamics from observational perturbations.

\begin{figure}[h]
    \centering
    \begin{tabular}{ccc}
        {(a)} & {(b)} & {(c)}  \\
        \includegraphics[height=3.5cm, width=0.33\linewidth]{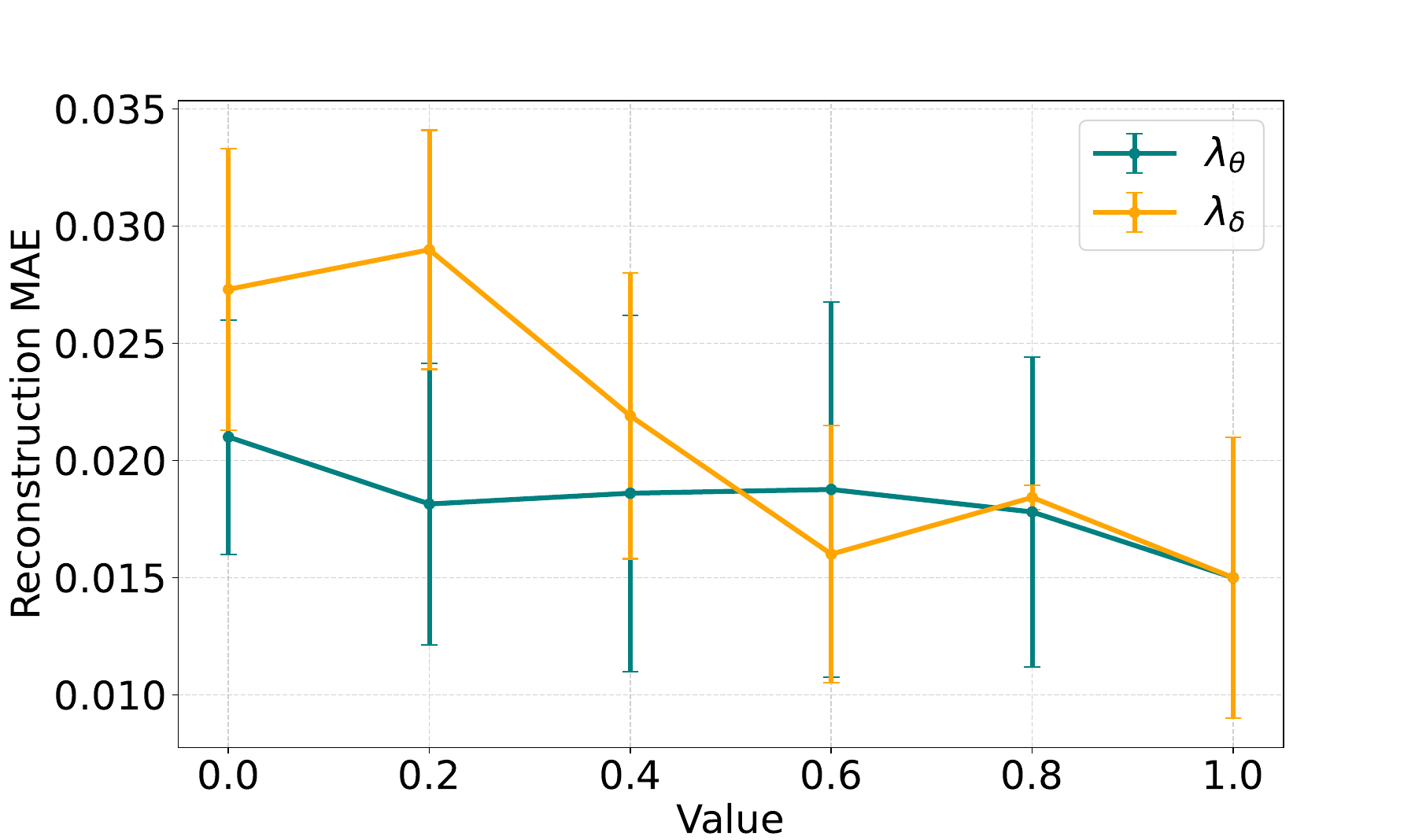} &
        \includegraphics[height=3.5cm, width=0.33\linewidth]{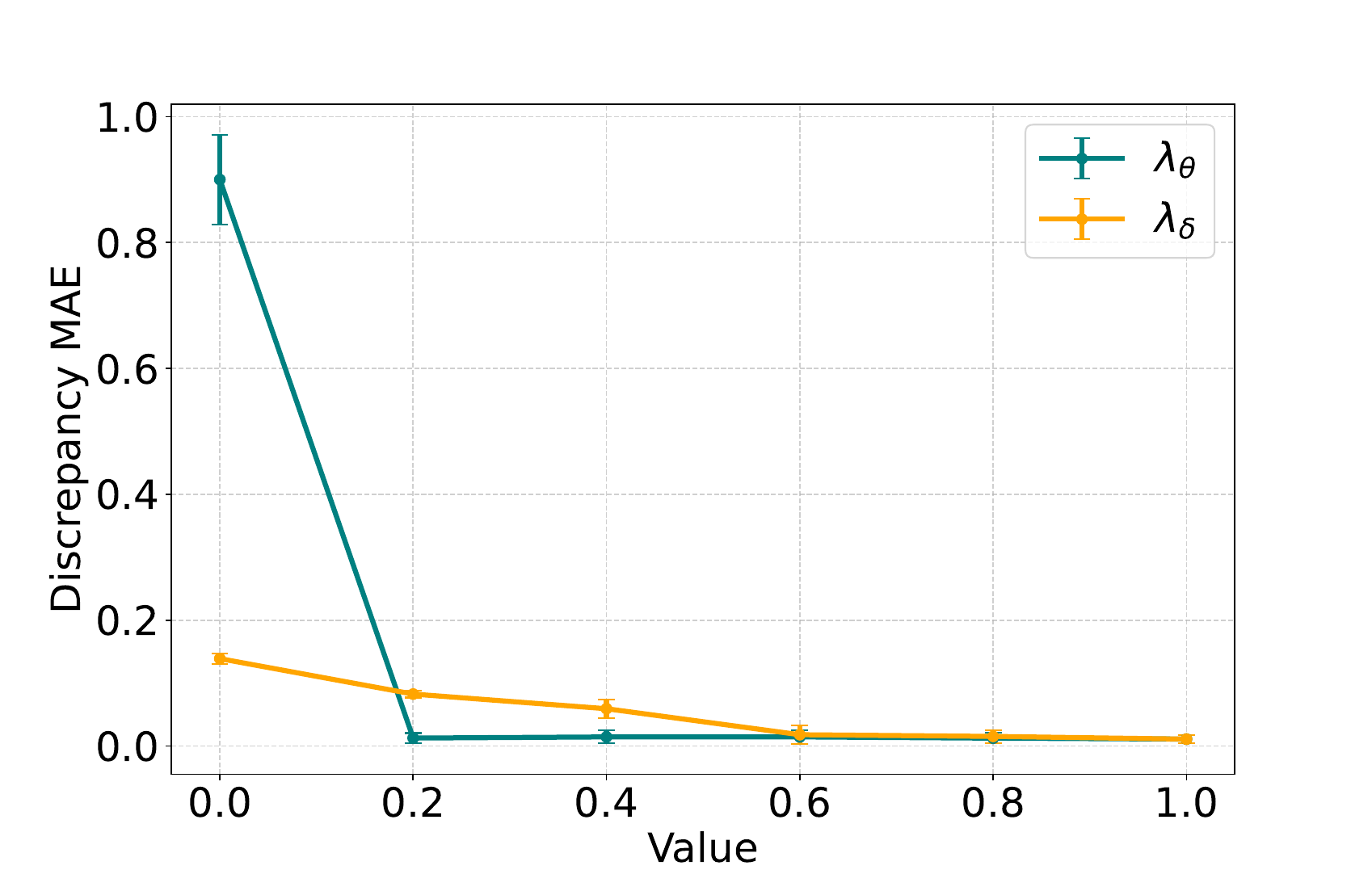} &
        \includegraphics[height=3.5cm, width=0.33\linewidth]{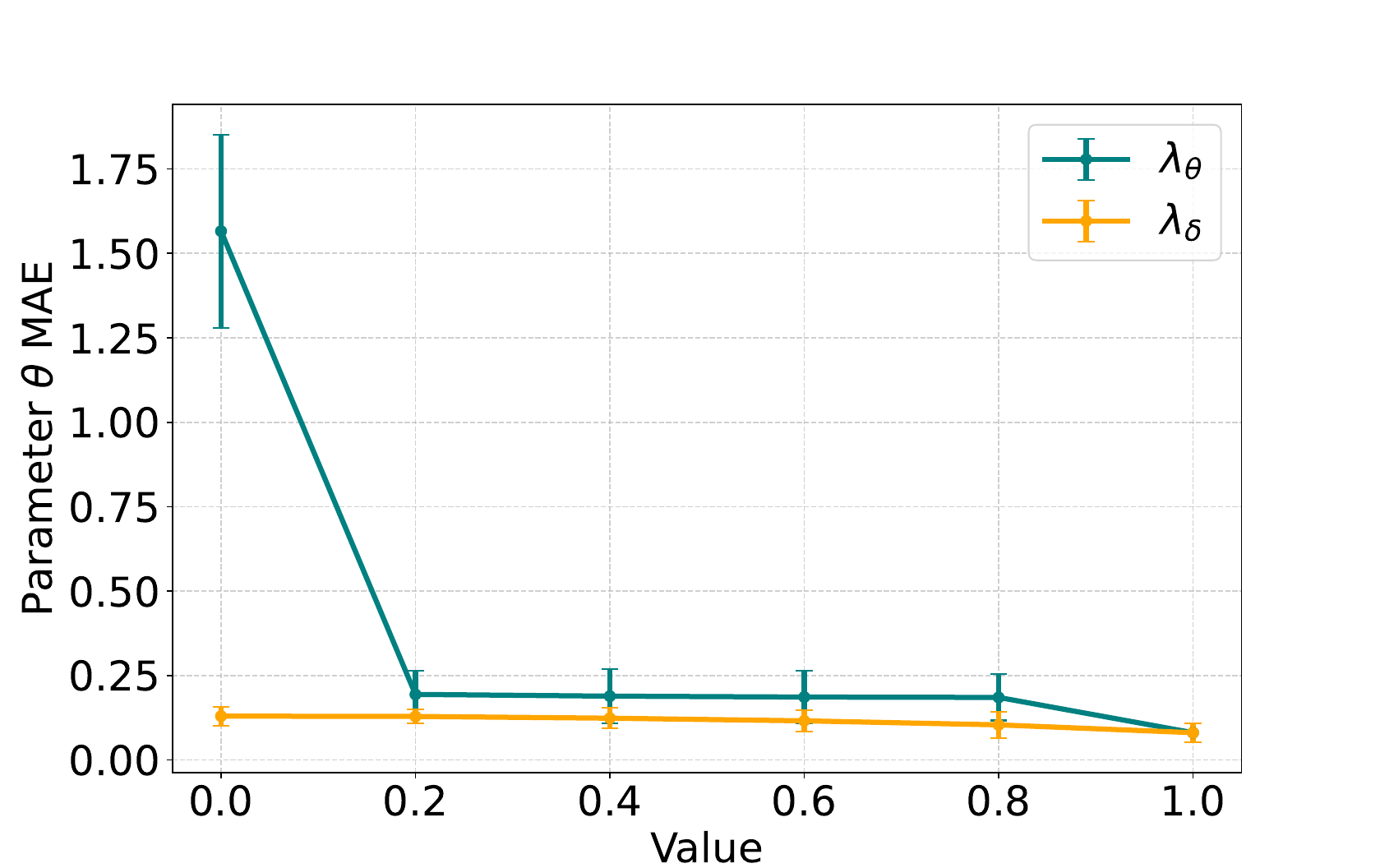}  \\
    \end{tabular}

    \caption{Ablation analysis on the influence of $\lambda_\theta$ (green) and $\lambda_\delta$ (yellow) on (a) reconstruction MAE, (b) Discrepancy MAE and (c) Physics parameter $\theta$ MAE. With increasing values of $\lambda_\theta$ and $\lambda_\delta$, the MAE values decrease. }
    \label{fig:ablation_lambda}
\end{figure}

\paragraph{Effect of $\lambda_\theta$ and $\lambda_\delta$.} 
We investigate the impact of the regularization parameters $\lambda_\theta$ (green curves) and $\lambda_\delta$ (yellow curves) in \autoref{fig:ablation_lambda} on three key metrics: reconstruction MAE, discrepancy MAE, and physics parameter $\theta$ MAE. Each metric is visualized with two curves: one showing the effect of varying $\lambda_\theta$ while keeping $\lambda_\delta=1$, and the other showing the effect of varying $\lambda_\delta$ with $\lambda_\theta=1$.

Reconstruction MAE \autoref{fig:ablation_lambda} (a): We observe that reconstruction error remains relatively stable across the range of $\lambda_\theta$ and $\lambda_\delta$, with a slight decrease as either parameter increases. Notably, the reduction in reconstruction MAE is more pronounced when increasing $\lambda_\theta$, suggesting that emphasizing the physics parameter regularization slightly improves the reconstruction quality.

Discrepancy MAE \autoref{fig:ablation_lambda} (b): The discrepancy MAE exhibits a stronger dependence on the regularization parameters. When $\lambda_\theta=0$, the error is initially high, but increases in $\lambda_\theta$ result in a sharp drop in MAE, particularly at $\lambda_\theta=0.2$, after which the error gradually saturates with a modest decrease for larger values. In contrast, increasing $\lambda_\delta$ leads to a smoother, more gradual reduction in discrepancy MAE, indicating a consistent improvement in aligning predictions with observations as the discrepancy regularization is strengthened.

Physics Parameter $\theta$ MAE \autoref{fig:ablation_lambda} (c): The trend for parameter MAE closely mirrors that of discrepancy MAE. Increasing $\lambda_\theta$ produces a rapid initial reduction in error followed by saturation, while higher $\lambda_\delta$ values yield a steady, gradual decline in MAE. This suggests that both regularization terms contribute to better estimation of the underlying physics parameters, with $\lambda_\theta$ having a more immediate impact and $\lambda_\delta$ providing stable long-term improvements.

\paragraph{Regularization methods.} In this study we evaluate the different regularization strategies $\mathcal{R}$ to obtain $\mathcal{L}_\delta$. $L_1$ and $L_2$ regularization penalize large discrepancies, encouraging sparsity ($L_1$) or smooth, small-magnitude discrepancies ($L_2$). Orthogonal regularization encourages the discrepancy network to be orthogonal to the physics latent, promoting disentanglement. This is given as $||\mathbf{z}^{(\delta)T}.\mathbf{z}^{(\theta)}||^2$. It encourages the discrepancy network to represent residual variation that is independent of the physics latent, thereby promoting disentanglement. Spectral weight regularization~\citep{venkataramanan2025distance} bounds the largest and smallest singular values of the weight matrices to prevent the discrepancy network from dominating the physics contribution. Gradient penalty~\citep{gulrajani2017improved} discourages large gradients of the discrepancy with respect to the inputs, promoting smooth outputs. As shown in Table~\ref{tab:regularizer_ablation}, L2 and spectral weight regularization achieve the best balance, producing low reconstruction and discrepancy errors while preserving accurate recovery of the physics parameters. Orthogonal regularization can enforce disentanglement but may reduce the accuracy of physics parameters, whereas gradient penalties overly constrain the network, resulting in higher errors. These results suggest that moderate capacity constraints are more effective than strict disentanglement or smoothness penalties for encouraging separation of physics and discrepancy roles.

\begin{table}[t]
\caption{Ablation study on the different regularization methods for the discrepancy network ($\delta$). }
    \begin{adjustbox}{width=0.6\textwidth, center}
    \centering
    \begin{tabular}{cccc}
    \toprule
     Regularization & Reconstruction MAE & $\delta$ MAE & $\theta$ MAE \\\midrule
     L1 & 0.045$\pm$0.017 & 0.077$\pm$0.034 & 0.204$\pm$0.106 \\
     L2 & \textbf{0.015$\pm$0.006} & \textbf{0.011$\pm$0.006} & \textbf{0.081$\pm$0.028} \\
     Orthogonal reg. & 0.029$\pm$0.009 & 0.023$\pm$0.019 & 0.198$\pm$0.086\\
     Spectral weight reg. & \underline{0.021$\pm$0.008} & \underline{0.015$\pm$0.010} & \underline{0.121$\pm$0.099}\\
     Gradient penalty & 0.037$\pm$0.002 & 0.084$\pm$0.015 & 0.244$\pm$0.147\\

    \bottomrule
    \end{tabular}
    \label{tab:regularizer_ablation}
    \end{adjustbox}
\end{table}

\paragraph{Effect of number of trajectory sample.}
This experiment studies how APIC performance scales with available data for the damped spring system. The results are shown in Table~\ref{tab:traj_sample}. Performance improves with more training realizations but begins to plateau beyond 64,000, suggesting that the gains are not simply due to data volume but reflect the amortized inference framework learning shared structure across realizations.

\begin{table}[]
    \centering
    \caption{Ablation study on the number of trajectory samples.}
    \begin{tabular}{cccc} \toprule
        N & Recon MAE & Param MAE & LL  \\ \midrule
        8000 & 0.033$\pm$0.015 & 1.263$\pm$0.027 & 0.424$\pm$ 0.019 \\
        16,000 & 0.019$\pm$0.010 & 0.625$\pm$0.032 & 0.778$\pm$0.017\\
        32,000 & 0.019$\pm$0.008  & 0.241$\pm$0.024 & 0.751$\pm$0.018\\
        64,000 & 0.017$\pm$0.008 &  0.118$\pm$0.028 & 0.883$\pm$0.020\\
        96,000 & 0.016$\pm$0.006 & 0.102$\pm$0.022 & 0.966$\pm$0.019\\
        128,000 & 0.015$\pm$0.006 & 0.081$\pm$0.028 & 1.121$\pm$0.017 \\ \bottomrule
    \end{tabular}
    \label{tab:traj_sample}
\end{table}

\section{Limitations and Future Work}
\label{sec:ap_lim}
Our current formulation assumes that the nominal simulator is a white-box, differentiable end-to-end model, allowing gradients to flow through the physics solver during training and calibration. This assumption is increasingly reasonable in modern hybrid modeling, where differentiable programming and neural surrogates are commonly used. Still, many high-fidelity industrial simulators remain black-box (e.g., legacy codes, proprietary solvers, or pipelines with non-differentiable components), which limits the direct applicability of our approach. A natural direction is to extend the framework to black-box simulators, for example, via gradient-free calibration, implicit differentiation with adjoints when available, surrogate-assisted optimization, or likelihood-free/simulation-based inference.

We assume that phase-1 training can access many simulator samples, implicitly treating simulation as cheap. This is a limitation for settings with expensive forward solves, large parameter spaces, or restricted access to simulation. Future work should address efficient learning under limited simulation budgets, including active task selection, multifidelity training, and the use of pretrained foundation surrogates. In particularly constrained regimes, amortized “zero-shot” calibration may effectively become a few-shot adaptation problem, where a pretrained calibrator is fine-tuned on a small number of observations per task. A principled treatment of this continuum from zero-shot to few-shot to per-task optimization would clarify when amortization is beneficial and how to allocate compute between pretraining and adaptation.

Like classic Kennedy–O’Hagan-style approaches, we treat the discrepancy as global, i.e., potentially present everywhere in the domain. In many scientific systems, misspecification is localized: confined to a subregion, boundary segment, time window, or regime (e.g., turbulence regions, contact events, switching dynamics). A promising extension is to incorporate localized discrepancy priors or gating mechanisms, allowing expert knowledge to specify candidate subdomains where discrepancy can act, or to automatically learn sparse/compact supports. This could improve identifiability and yield more interpretable corrections.


We allow different structural forms for the correction, additive or multiplicative, depending on the domain. This choice is consequential: it shapes both the learned correction and its interpretation. For example, in the spring system, a multiplicative envelope is natural, whereas in other settings, an additive forcing term is more appropriate. However, when the true misspecification is not well matched to the chosen form, the learned discrepancy can become harder to interpret. Future work could study model selection over discrepancy parameterizations, mixtures of additive/multiplicative forms, or physics-guided parameterizations that encode conservation laws and invariances.

\end{document}